\documentclass[conference]{IEEEtran}
\IEEEoverridecommandlockouts

\usepackage{diagbox}
\usepackage{cite}
\usepackage{amsmath,amssymb,amsfonts}
\usepackage{algorithmic}
\usepackage{graphicx}
\usepackage{textcomp}
\usepackage{xcolor}
\usepackage{subcaption}
\usepackage{booktabs}
\usepackage{comment}
\usepackage{url}
\usepackage{multirow}
\def\BibTeX{{\rm B\kern-.05em{\sc i\kern-.025em b}\kern-.08em
    T\kern-.1667em\lower.7ex\hbox{E}\kern-.125emX}}
\usepackage{makecell}
\usepackage{colortbl}
\usepackage{tikz}
\usetikzlibrary{positioning, arrows.meta, shapes.geometric}    

\definecolor{siggreen}{RGB}{185, 230, 195}
\definecolor{medgreen}{RGB}{215, 240, 220}
\definecolor{lightgreen}{RGB}{235, 248, 237}
\definecolor{darkred}{RGB}{160, 40, 40}
\definecolor{sigred}{RGB}{235, 150, 150}      
\definecolor{lightred}{RGB}{248, 210, 210}    
\newcommand{\rev}[1]{\textcolor{blue}{#1}}
\newcommand{\CL}{\cellcolor{siggreen}\checkmark\ large}
\newcommand{\CM}{\cellcolor{medgreen}\checkmark\ med.}
\newcommand{\CS}{\cellcolor{lightgreen}\checkmark\ small}
\newcommand{\NSs}{\cellcolor{lightred}\textcolor{darkred}{$\times$\ small}}
\newcommand{\NN}{\cellcolor{sigred}\textcolor{darkred}{$\times$\ negl.}}

\makeatletter
\newcommand{\linebreakand}{%
  \end{@IEEEauthorhalign}
  \hfill\mbox{}\par
  \mbox{}\hfill\begin{@IEEEauthorhalign}
}
\makeatother

\begin{document}

\title{Bi-temporal Image-driven Acute Stroke Evolution Analysis

}

\author{
\IEEEauthorblockN{Md Sazidur Rahman\textsuperscript{1},
Kjersti Engan\textsuperscript{1},
Kathinka D{\ae}hli Kurz\textsuperscript{1,2},
Mahdieh Khanmohammadi\textsuperscript{1}}
\IEEEauthorblockA{\textsuperscript{1}\textit{Department of Electrical Engineering and Computer Science, University of Stavanger}, Stavanger, Norway\\
\textsuperscript{2}\textit{Stavanger Medical Imaging Laboratory (SMIL), Department of Radiology, Stavanger University Hospital}, Stavanger, Norway\\
\{md.s.rahman, kjersti.engan, mahdieh.khanmohammadi\}@uis.no, kathinka.dehli.kurz@sus.no}
}

\maketitle
{\renewcommand{\thefootnote}{}%
\footnotetext{\copyright~2026 IEEE. Personal use of this material is permitted. Permission from IEEE must be obtained for all other uses, in any current or future media, including reprinting/republishing this material for advertising or promotional purposes, creating new collective works, for resale or redistribution to servers or lists, or reuse of any copyrighted component of this work in other works.}}
\begin{abstract}
Acute ischemic stroke requires rapid treatment decisions that are strongly guided by emergency imaging. Admission computed tomography perfusion (CTP) is commonly used to estimate the ischemic core, representing irreversibly damaged tissue, and the penumbra, representing hypoperfused but potentially salvageable tissue. Follow-up diffusion-weighted MRI (DWI) is then used to define the final infarct. Existing imaging approaches primarily focus on core--penumbra segmentation or final-infarct prediction, but provide limited insight into how heterogeneous penumbral tissue evolves after treatment. We propose a bi-temporal tissue phenotyping framework that links admission CTP signatures with follow-up DWI-defined tissue outcome using six outcome-aware region-of-interest classes. Admission tissue signatures are characterized using statistical, radiomic, and deep-learning features extracted from mJ-Net and nnU-Net representations. On an internal cohort (SUH), salvaged and infarcted penumbra showed consistent feature-space separation ($\tilde{\Delta}{\cos}=0.146$, $p<0.05$), while core tissue showed minimal separation by subsequent fate. The largest separation was observed between initially non-hypoperfused tissue that later infarcted and healthy contralateral tissue ($\tilde{\Delta}{\cos}=0.460$, $p<0.05$). Cross-dataset evaluation on the  publicly available ISLES'24 dataset showed similar trends, supporting the consistency of the observed feature-space patterns.
These findings suggest that admission CTP contains outcome-associated tissue information beyond conventional core-penumbra delineation. The code is available on \rev{GitHub}\footnote{\begin{tabular}[t]{@{}l@{}}
\rev{Code: \url{https://github.com/yokko123/bi-temporal-ctp-dwi-code}}
\end{tabular}}.
\end{abstract}


\section{Introduction}
Acute ischemic stroke is a major cause of disability and requires rapid assessment of brain tissue viability to guide treatment decisions \cite{feiginWorldStrokeOrganization2025}. Computed tomography perfusion (CTP), routinely acquired at hospital admission, is used to estimate ischemic core and penumbra representing irreversibly damaged and potentially salvageable tissue \cite{ctp_wintermark}. However, tissue outcome is not fixed at admission, and regions with similar initial appearance may later evolve differently following treatment. Stroke progression therefore represents a temporal process that cannot be fully described from admission imaging alone.

Follow-up diffusion-weighted MRI (DWI) provides the final infarct and is commonly used to evaluate tissue outcome \cite{yang_agreement_2022}. Many automated approaches analyze CTP acquired at admission to estimate ischemic core and penumbra using thresholding or machine learning-based segmentation of perfusion parametric maps \cite{tomasetti_machine_2022,tomasetti_multi-input_2022,Freda}. More recent methods leverage the raw spatiotemporal CTP signal for direct segmentation of ischemic regions \cite{tomasetti2023ct}, while others aim to predict final infarct outcome from admission imaging \cite{amador2022predicting}. Although these approaches improve automated delineation and outcome prediction, they primarily formulate the problem as segmentation or prediction and treat ischemic regions as homogeneous targets. Consequently, they provide limited insight into how penumbral tissue with similar admission-time appearance evolves toward recovery or infarction.
\begin{figure*}[htbp]
  \centering
  \includegraphics[width=0.86\textwidth]{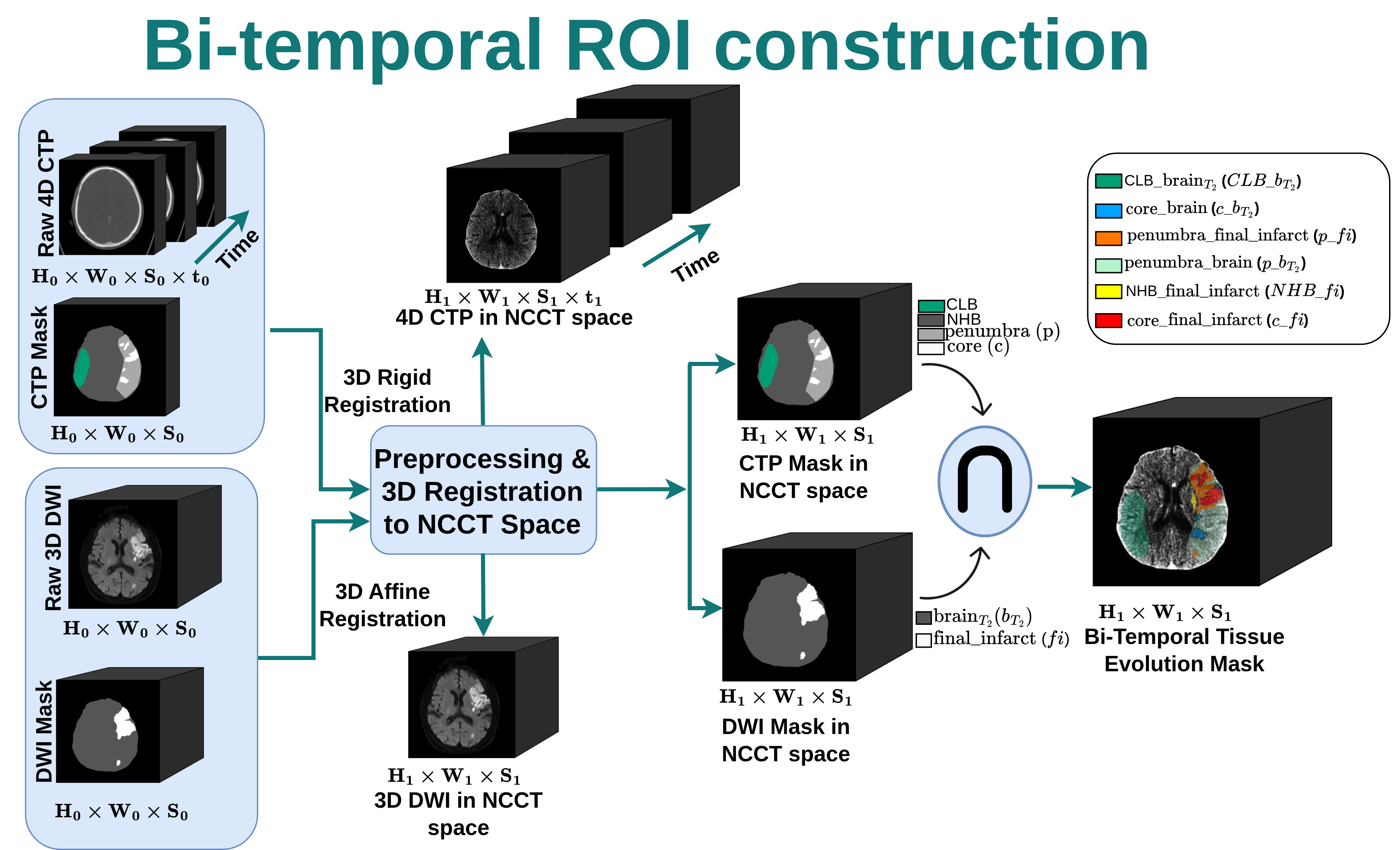}

  \vspace{0.3em}

  \includegraphics[width=0.9\textwidth]{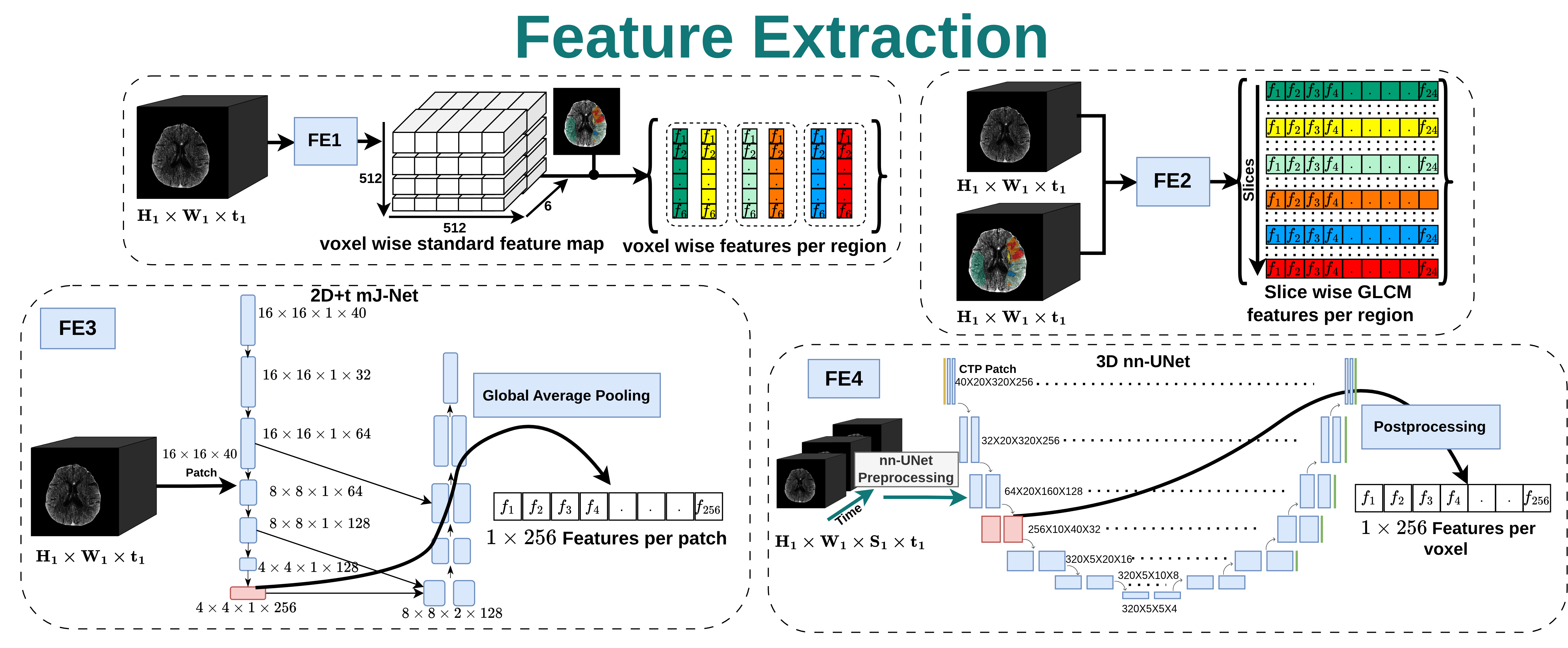}
\caption{Framework Overview. (\textbf{Top}): bi-temporal ROI construction, intersecting registered admission CTP and follow-up DWI masks into six outcome-aware classes. (\textbf{Bottom}): CTP feature extraction (FE1--FE4). $\text{ROI}_x^{y}$, admission class $x$ and outcome $y$: $c$ core, $p$ penumbra, NHB non-hypoperfused, CLB contralateral, $b_{T_2}$ non-infarcted, $fi$ final infarct; NCCT non-contrast CT.}
  \label{fig:framework}
\end{figure*}

Recent advances in radiomics for stroke imaging \cite{jiang2023radiomics,vantimmerenRadiomicsMedicalImaginghowto2020} and deep embeddings from CT imaging \cite{tran2025comparing} enable extraction of quantitative image representations beyond conventional perfusion parameters. Such representations make it possible to characterize tissue properties directly from imaging data. Imaging phenotyping aims to identify data-driven tissue subtypes that reflect different tissue states \cite{stroke_pheno1}. 
In stroke imaging, such phenotyping has so far largely characterized tissue at a single time point~\cite{stroke_pheno1}, without explicitly linking admission signatures to eventual tissue fate. A bi-temporal view, connecting admission CTP to follow-up DWI outcome and studying tissue evolution over time, therefore remains comparatively unexplored, and motivates the framework proposed here.

In this work, we present a feasibility study for ischemic tissue evolution by linking admission-time perfusion signatures to follow-up tissue outcome through a bi-temporal image-driven tissue phenotyping framework, illustrated in Fig.~\ref{fig:framework}. CTP acquired at admission ($T_1$) and DWI obtained at follow-up ($T_2$) are spatially aligned to construct outcome-aware regions describing tissue recovery and infarction. Admission tissue is characterized using statistical descriptors, radiomic features, and deep convolutional neural network embeddings, enabling quantitative analysis of tissue organization and evolution in feature space.

We hypothesize that ischemic tissue progression can be described as distinct tissue phenotypes in feature space, where admission-time imaging signatures encode latent tissue states associated with recovery or infarction. Identifying such phenotypes provides a richer characterization of ischemic tissue vulnerability beyond conventional perfusion thresholds, with potential relevance for understanding treatment response and infarct growth risk. Penumbral fate is also strongly influenced by post-admission reperfusion. We therefore do not claim that admission CTP determines outcome; however, knowledge of the complexity of the ischemic region may eventually lead to information that can be valuable for clinicians in the acute setting for therapy decisions.

\section{Materials \& Methods}
\subsection{Internal Cohort Description:}
This retrospective study used data from Stavanger University Hospital (SUH), Norway (approved by Regional Committee for Medical and Health Research Ethics (REK) under Project 2012/1499) acquired at admission ($T_1$), follow-up ($T_2$), and three-month outcome assessment. At $T_1$, patients underwent the full CT protocol with clinical assessment including NIHSS and baseline mRS; the raw CTP sequence comprised 30 temporal frames acquired at non-uniform intervals (1\,s for the first 20 frames, 2\,s for the final 10). At $T_2$, typically 24--72\,hours after stroke onset, MRI including DWI and ADC was acquired with repeat NIHSS assessment. Functional outcome was assessed at three months using mRS.

The full cohort included 149 patients scanned between January 2014 and August 2020: 74 large vessel occlusion (LVO), 60 non-LVO ischemic stroke, and 15 without ischemic stroke (WIS). The mean age was $71.9 \pm 14.4$ years, and LVO patients had higher baseline NIHSS than non-LVO patients ($13.2 \pm 9.0$ vs.\ $6.2 \pm 5.2$). All LVO and non-LVO patients received intravenous thrombolysis, and $42$ ($28.2\%$) patients  additionally underwent mechanical thrombectomy. Patients with confirmed ischemic stroke, complete CTP and DWI, manual annotations, and successful registration formed the final analysis cohort 
($n=109$; LVO $=63$, non-LVO $=46$; Table~\ref{tab:cohorts}). The remaining $40$ patients were excluded from the bi-temporal analysis only; all $149$ were used for CNN model development (Section~\ref{sec:impl}).

\begin{table}[t]
\centering
\caption{Summary of the SUH and ISLES'24 cohorts.}
\label{tab:cohorts}
\footnotesize
\setlength{\tabcolsep}{4pt}
\renewcommand{\arraystretch}{1.1}
\begin{tabular}{@{}lll@{}}
\toprule
\textbf{Characteristic} & \textbf{SUH (after preproc.)} & \textbf{ISLES'24} \\
\midrule
Patients (analysis) & 109/149 & 149/149 \\
Age, y (mean$\pm$SD) & 71.9$\pm$14.4 & 71.1$\pm$14.7 \\
Admission NIHSS & 9.4$\pm$8.3 & 12.0$\pm$6.3 \\
Imaging ($T_1$$\rightarrow$$T_2$) & CT $\rightarrow$ MRI & CT $\rightarrow$ MRI \\
Median Size (NCCT grid) & $512 \times 512 \times 50$ & $512 \times 581 \times 69$ \\
CTP temporal res. & 1\,s & 1\,s \\
Voxel resolution (NCCT grid) & 0.41 $\pm$ 0.02 mm & 0.41 $\pm$ 0.05 mm \\
Slice thickness (NCCT grid) & 3 mm & 2.54 $\pm$ 0.90 mm\\
Follow-up interval & 24--72\,h & 2--9\,d \\
$T_1$ core/penumbra annotation &  ground truth & automated (nnU-Net) \\
$T_2$ final infarct annotation & ground truth & ground truth \\
\bottomrule
\end{tabular}
\end{table}

\subsection{ISLES'24 Cohort Description}
To assess cross-dataset consistency, we include the publicly available ISLES'24 training set~\cite{riedelIschemicStrokeLesion2026}, comprising $n=149$ cases from two centers: the University Hospital of the Technical University of Munich ($n=99$) and the University Hospital of Zurich ($n=50$). All patients underwent acute CT imaging (NCCT, CT angiography (CTA), CTP with derived perfusion maps) followed by intracranial reperfusion therapy, restricted to near-complete or complete recanalization (mTICI 2c/3). CTP has a temporal resolution of 1\,s/frame. Follow-up MRI (DWI and ADC) was acquired 2--9 days after admission. The cohort had a mean age of $71.1 \pm 14.7$ years and admission NIHSS of $12.0 \pm 6.3$. The preprocessed CTP and DWI, co-registered to the NCCT space were used directly without additional registration. Final infarct masks at $T_2$ were produced via a hybrid human-algorithm approach~\cite{riedelIschemicStrokeLesion2026}.  

\textbf{Ground Truth and Annotation:}
For the SUH cohort, expert neuroradiologists annotated tissue regions using a MATLAB-based tool following a previously described protocol~\cite{tomasetti2023ct, tomasetti_multi-input_2022}. At $T_1$, CTP-derived perfusion maps (cerebral blood flow (CBF), cerebral blood volume (CBV), time-to-maximum (Tmax)) and maximum intensity projection (MIP) images were reviewed: hypoperfused regions with prolonged Tmax and reduced CBF but preserved CBV were labeled penumbra ($p$), additional CBV reduction indicated ischemic core ($c$), and the remainder non-ischemic brain ($b$). At $T_2$, final infarct ($fi$) was defined by cytotoxic edema on DWI. For ISLES'24, which provides no manual $T_1$ annotations, core and penumbra were obtained automatically using an nnU-Net~\cite{isensee2021nnu} trained on the SUH cohort, while the $T_2$ final infarct was taken from the dataset's hybrid human-algorithm masks~\cite{riedelIschemicStrokeLesion2026}.

\textbf{Preprocessing and Registration:} For the SUH cohort, all imaging data (4D CTP, NCCT, DWI, and ADC) were first converted from DICOM to NIfTI. The 4D CTP was motion-corrected by rigidly registering each temporal frame to the first frame, and the final 10 frames were temporally resampled to a uniform 1\,s rate by linear interpolation, yielding a 40-frame sequence. All volumes were then co-registered to a common reference frame defined by the admission NCCT, establishing voxel-level correspondence across modalities and time points: the motion-corrected CTP was aligned to the NCCT by 3D rigid registration (SimpleITK~\cite{lowekampDesignSimpleITK2013}) and the follow-up DWI by 3D affine registration~\cite{lowekampDesignSimpleITK2013}. The estimated transforms were propagated to all 40 CTP frames, the DWI volume, and their annotation masks, with the CTP resampled from 5\,mm to 3\,mm slice thickness. Brain tissue was extracted using SynthStrip~\cite{hoopesSynthStripSkullstrippingAny2022}. Lastly, CTP intensities were windowed to $[0,100]$\,HU and rescaled to an 8-bit gray-level range prior to feature extraction for both SUH and ISLES'24 cohorts.

\subsection{Bi-temporal Region of Interest (ROI) Construction}
We denote the preprocessed \& registered CTP as $V(v,t)$, with $v=(x,y,z)\in\Omega\subset\mathbb{Z}^3$ the volumetric index and $t\in\{1,\dots,40\}$ the time index. For a fixed axial level $z_i$, the spatiotemporal slice is $V_{z_i}(\bar{x})$, where $\bar{x}=(x,y,t)$ indexes the two spatial dimensions and time. The follow-up DWI is denoted $D(v)$, with spatial slice $D_{z_i}(\tilde{x})$ and $\tilde{x}=(x,y)$. The registered DWI--CTP pair is then $(D(v),V(v,t))$ volumetrically and $(D_{z_i}(\tilde{x}),V_{z_i}(\bar{x}))$ for a single slice.
\begin{table}[htbp]
\centering
\caption{Median voxels per patient [IQR] by ROI class.}
\label{tab:voxels_per_patient}

\begin{tabular}{lcc}
\toprule
\textbf{ROI class} & \textbf{SUH (n=109)} & \textbf{ISLES'24 (n=149)} \\
\midrule
\textbf{core\_fi}    & 1{,}451 [212--7{,}832]         & 2{,}450 [347--8{,}444]           \\
\textbf{core\_brain} & 1{,}569 [268--5{,}933]         & 977 [286--4{,}476]               \\
\textbf{pen\_fi}     & 6{,}924 [996--17{,}953]        & 11{,}594 [4{,}715--41{,}232]     \\
\textbf{pen\_brain}  & 70{,}189 [17{,}834--155{,}738] & 170{,}273 [91{,}339--279{,}119]  \\
\textbf{clb\_brain}  & 467{,}646 [251{,}024--647{,}513] & 934{,}954 [763{,}774--1{,}060{,}942] \\
\textbf{NHB\_fi}     & 3{,}216 [851--8{,}992]         & 9{,}792 [3{,}452--27{,}560]      \\
\bottomrule
\end{tabular}

\end{table}
Six bi-temporal ROIs were constructed for both cohorts by intersecting the $T_1$ tissue and $T_2$ outcome labels described above. For each pair $(R_{T_1}, R_{T_2})$, we define $\text{ROI}_{R_{T_1}}^{R_{T_2}} := R_{T_1} \cap R_{T_2}$, forming six classes: $\text{ROI}_{c}^{fi}$ (core, remained infarcted); $\text{ROI}_{c}^{b}$ (core, salvaged); $\text{ROI}_{p}^{fi}$ (penumbra, progressed to infarct); $\text{ROI}_{p}^{b}$ (penumbra, salvaged); $\text{ROI}_{CLB}^{b}$ (contralateral healthy brain, remained healthy); $\text{ROI}_{NHB}^{fi}$ (non-hypoperfused brain, later infarcted). To ensure spatial consistency, the core mask was dilated and subtracted from the penumbra mask, and the penumbra mask was similarly dilated and subtracted from the remaining brain mask. The contralateral hemisphere and ventricular regions were defined using SynthSeg~\cite{billotSynthSegSegmentationBrain2023}; ventricular voxels were excluded from all analyses. Voxels not assigned to any class were treated as background and excluded. 
Table~\ref{tab:voxels_per_patient} reports the median number of voxels per patient in each ROI class for both cohorts, indicating the relative size of each class and the variability across patients.

\subsection{Feature Extraction}
\label{sec:FE}
To characterize ischemic tissue, we perform feature extraction~(FE) from preprocessed CTP and relate them to final outcome on DWI. We consider three feature families: baseline statistical descriptors, radiomic texture metrics, and deep CNN embeddings to investigate tissue heterogeneity and provide a compact representation of stroke evolution.

\textbf{FE1: Baseline (BL) features} (Fig.~\ref{fig:framework}) are directly
derived from CTP, $V_{z_i}(\bar{x})$ using a sliding window
$\mathcal{W}(\bar{x}) \in \mathbb{R}^{3\times3\times40}$ centered around $\bar{x}$.
We compute six first-order statistics: mean, standard deviation, skewness,
kurtosis, minimum and maximum. Let $g:\mathbb{R}^{3\times3\times40}\to\mathbb{R}^6$
denote the feature operator. The baseline feature vector for each slice and
location $\bar{x}$ is
$\boldsymbol{f}^{\mathrm{BL}}_{z_i}(\bar{x})=g\big(V_{z_i}(\mathcal{W}(\bar{x}))\big)\in\mathbb{R}^6$.
Location-wise feature vectors are mapped to bi-temporal tissue evolution masks
($\text{ROI}_{R_{T_1}}^{R_{T_2}}$). For each patient $pt$, slice $z_i$ and ROI, a
single descriptor $\boldsymbol{F}^{\mathrm{BL}}_{pt,\mathrm{ROI},z_i}\in\mathbb{R}^6$
is obtained by max-pooling aggregation, such that each tissue-evolution class and
slice is represented by one feature vector. Slice-level descriptors are further
aggregated to a patient-level representation
$\boldsymbol{F}^{\mathrm{BL}}_{pt,\mathrm{ROI}}\in\mathbb{R}^6$ by max-pooling
across slices $z_i$.

\textbf{FE2: Radiomic features} (Fig.~\ref{fig:framework}) capture higher-order co-occurrence patterns beyond first-order statistics. For each axial slice, the CTP sequence is treated as a 3D spatiotemporal volume, with two spatial dimensions and time as the third dimension, $V_{z_i}(\bar{x})$ where $\bar{x} = (x, y, t)$. Using \texttt{PyRadiomics}~\cite{pyradiomics}, we extract 24 symmetrical GLCM features~\cite{glcm} from each slice-wise spatiotemporal volume paired with an ROI mask stacked across all time points. Based on Mann--Whitney U tests~\cite{mann_test_1947}, FDR correction~\cite{benjaminiControllingFalseDiscovery1995}, and Cliff's delta~\cite{cliff_dominance_1993}, four features were retained: Imc1, Imc2, MCC, and Correlation. These form $\boldsymbol{F}^{\mathrm{GLCM}}_{pt,\mathrm{ROI},z_i}\in\mathbb{R}^4$ and are aggregated to patient level by max-pooling across slices.

\textbf{FE3 \& FE4: Deep CNN-based embeddings} (Fig.~\ref{fig:framework}) capture hierarchical, data-driven tissue representations, extracted in inference mode from two CTP segmentation networks: (i)~temporal-aware 2D+time mJ-Net~\cite{tomasetti2020cnn}, designed for core and penumbra segmentation from CTP; and (ii)~3D nnU-Net~\cite{isensee2021nnu}, which stacks 40 time points as channels and serves as a widely-used segmentation backbone for comparison.

For mJ-Net, spatio-temporal patches ($16\times16\times40$, stride~1) are passed through the encoder; activations from the last convolutional block ($4\times4\times1\times256$) are globally average-pooled, yielding a 256-D vector per patch center. 

For nnU-Net, the CTP volume is tiled into overlapping 3D patches (50\% overlap), and stage-3 encoder features (256-D, matching mJ-Net) are extracted, upsampled, and stitched into a full-volume map $\mathcal{F}(v)\in\mathbb{R}^{256}$ resampled to NCCT space. For both models, $\boldsymbol{F}^{\mathrm{CNN}}_{pt,\mathrm{ROI},z_i}\in\mathbb{R}^{256}$ is obtained by max-pooling over ROI voxels per slice and then across slices $z_i$.

\begin{table}[t!]
\centering
\caption{Region-pair comparisons. FE1/FE2: Mann--Whitney U 
(FDR, $\alpha{=}0.05$), Cliff's $d$ effect size.
FE3/FE4: median $\tilde{\Delta}_{\cos}$, one-sample Wilcoxon.
Symbol\,=\,significance (\textbf{green}:\checkmark/ 
$\textbf{red}:\times$), shade\,=\,effect-size magnitude 
(Cliff's $d$ at $0.147/0.33/0.474$; 
$\tilde{\Delta}_{\cos}$ at $0.1/0.3$).}
\label{tab:stats_4tests}
\setlength{\tabcolsep}{2pt}
\renewcommand{\arraystretch}{0.60}
{\fontsize{5.5}{6.5}\selectfont
\resizebox{\linewidth}{!}{%
\begin{tabular}{l |
  >{\centering\arraybackslash}p{1.9cm} |
  >{\centering\arraybackslash}p{1.9cm} |
  >{\centering\arraybackslash}p{1.9cm} |
  >{\centering\arraybackslash}p{1.9cm}}
\toprule
& \textbf{Test 1} & \textbf{Test 2} 
& \textbf{Test 3} & \textbf{Test 4} \\
\textbf{Feature}
& \makecell{$\text{ROI}_{p}^{b}$ vs 
  $\text{ROI}_{p}^{fi}$}
& \makecell{$\text{ROI}_{c}^{b}$ vs 
  $\text{ROI}_{c}^{fi}$}
& \makecell{$\text{ROI}_{\textit{NHB}}^{fi}$ vs 
  $\text{ROI}_{p}^{fi}$}
& \makecell{$\text{ROI}_{\textit{NHB}}^{fi}$ vs 
  $\text{ROI}_{\textit{CLB}}^{b}$} \\
\midrule
\multicolumn{5}{l}{\textit{SUH cohort ($n=109$)}} \\
\specialrule{0.4pt}{1pt}{1pt}
\multicolumn{5}{l}{\textbf{FE1}} \\
Mean      & \CL & \NSs & \CS  & \CL \\
Std       & \CM & \NSs & \NSs & \CL \\
Skewness  & \CM & \NN  & \NN  & \CL \\
Min       & \CL & \NSs & \NSs & \CL \\
Max       & \CS & \NSs & \NN  & \CS \\
Kurtosis  & \CL & \NN  & \NN  & \CL \\
\specialrule{0.4pt}{1pt}{1pt}
\multicolumn{5}{l}{\textbf{FE2}} \\
Imc1        & \CL  & \NSs & \NN  & \CL  \\
Imc2        & \CM  & \NN  & \CS  & \CS  \\
Correlation & \CL  & \NN  & \CS  & \NSs \\
MCC         & \CL  & \NSs & \CS  & \CS  \\
\specialrule{0.4pt}{1pt}{1pt}
$\tilde{\Delta}_{\cos}$ (FE3) 
  & \cellcolor{medgreen}\checkmark\ 0.146 
  & \cellcolor{lightgreen}\checkmark\ 0.031 
  & \cellcolor{medgreen}\checkmark\ 0.127 
  & \cellcolor{siggreen}\checkmark\ 0.460 \\
$\tilde{\Delta}_{\cos}$ (FE4) 
  & \cellcolor{lightgreen}\checkmark\ 0.091 
  & \cellcolor{lightgreen}\checkmark\ 0.038 
  & \cellcolor{lightgreen}\checkmark\ 0.090 
  & \cellcolor{medgreen}\checkmark\ 0.193 \\
\midrule
\multicolumn{5}{l}{\textit{ISLES'24 cohort ($n=149$)}} \\
\specialrule{0.4pt}{1pt}{1pt}
\multicolumn{5}{l}{\textbf{FE1}} \\
Mean      & \CL & \NN  & \CL  & \CL \\
Std       & \CL & \NN  & \CM  & \CL \\
Skewness  & \CM & \NN  & \NN  & \CL \\
Min       & \CL & \NN  & \CL  & \CL \\
Max       & \CS & \NN  & \CS  
  & \cellcolor{lightgreen}\checkmark\ negl. \\
Kurtosis  & \CL & \NN  & \CL  & \CL \\
\specialrule{0.4pt}{1pt}{1pt}
\multicolumn{5}{l}{\textbf{FE2}} \\
Imc1        & \CM  & \NN & \NN  & \CL \\
Imc2        & \CL  & \NN & \CL  & \CS \\
Correlation & \CM  & \NN & \CL  & \CS \\
MCC         & \CS  & \NN & \CL  & \NN \\
\specialrule{0.4pt}{1pt}{1pt}
$\tilde{\Delta}_{\cos}$ (FE3) 
  & \cellcolor{medgreen}\checkmark\ 0.177 
  & \cellcolor{lightgreen}\checkmark\ 0.044 
  & \cellcolor{medgreen}\checkmark\ 0.126 
  & \cellcolor{siggreen}\checkmark\ 0.401 \\
$\tilde{\Delta}_{\cos}$ (FE4) 
  & \cellcolor{medgreen}\checkmark\ 0.140 
  & \cellcolor{lightgreen}\checkmark\ 0.042 
  & \cellcolor{lightgreen}\checkmark\ 0.094 
  & \cellcolor{medgreen}\checkmark\ 0.115 \\
\bottomrule
\end{tabular}%
}}
\end{table}

\subsection{Assessment techniques}
Visualization tools including bubble plots and t-SNE plots~\cite{maatenVisualizingDataUsing2008} were used to qualitatively assess the feature space.
For each pair of bi-temporal ROIs, BL and GLCM features were compared using Mann-Whitney U tests~\cite{mann_test_1947} with FDR correction~\cite{benjaminiControllingFalseDiscovery1995} ($\alpha = 0.05$) and Cliff's delta $(d)$ effect sizes~\cite{cliff_dominance_1993}.
CNN features were z-score normalized and their similarities were quantified using absolute cosine similarity.
Given two max-pool aggregated slice-level embeddings $\bar{e}_i$ and $\bar{e}_j$, cosine similarity is defined as
$\operatorname{sim}(\bar{e}_i,\bar{e}_j) = \dfrac{|\bar{e}_i \cdot \bar{e}_j|}{\|\bar{e}_i\|\,\|\bar{e}_j\|}$.\\
The mean similarity between two sets of embeddings $R_1$ and $R_2$ is defined as:
\[\overline{\operatorname{sim}}(R_1,R_2)
=
\frac{1}{|R_1||R_2|}
\sum_{i \in R_1}
\sum_{j \in R_2}
\operatorname{sim}(\bar e_i, \bar e_j) \]
Within- and between-group similarities were computed using slice-aggregated CNN embeddings 
$\boldsymbol{F}^{\mathrm{CNN}}_{pt,\mathrm{ROI},z_i}$ for each patient, $pt$.
A separation index, $\Delta_{\cos}$, was defined to quantify embedding-space separation by final tissue state 
and was computed as the difference between average within-group similarity and between-group similarity within the same admission-defined tissue class (R is short for ROI):
\[
\begin{aligned}
\Delta_{\cos}{(pt)}
=
\frac{1}{2}
\Big[
\overline{\operatorname{sim}}\left(R_1^{(pt)}, R_1^{(pt)}\right)
+
\overline{\operatorname{sim}}\left(R_2^{(pt)}, R_2^{(pt)}\right)
\Big]\\
-
\overline{\operatorname{sim}}\left(R_1^{(pt)}, R_2^{(pt)}\right)
\end{aligned}
\]

Higher values of $\Delta_{\cos}(pt)$ indicate stronger embedding separation by final tissue state; all tables report its across-patient median, $\tilde{\Delta}_{\cos}$.
Statistical significance of the separation index was assessed using a one-sample Wilcoxon signed-rank test \cite{wilcoxon_individual_1945} against zero. Lower $\tilde{\Delta}_{\cos}$ indicates a small gap between within- and between-group cosine similarity, higher values a large one, visible as distinct t-SNE clusters; green shading in Tables~III--IV marks significant and red non-significant separation, with darker shades indicating larger effect sizes.

\section{Implementation Details}
\label{sec:impl}
The CNN feature extractors were developed using the full SUH CTP cohort ($n=149$) with a 116/33 train/test split and 5-fold cross-validation within the training set. After training, the resulting mJ-Net and nnU-Net models were used in inference mode to extract deep feature representations for ROI-level analysis. All experiments were implemented in Python using PyTorch, Numpy, Pandas and SciPy libraries.
\textbf{FE1 and FE2} feature extraction was performed on a 64-core CPU and 64 GB RAM. Baseline statistical features were computed using a sliding window of size $3\times3\times40$ with stride 1. GLCM features were extracted using \texttt{PyRadiomics}~\cite{pyradiomics} with symmetrical 3D GLCMs, producing 24 features per slice, of which four were retained based on discriminative capability. \textbf{FE3} was trained on NVIDIA A100 80\,GB GPUs for 100 epochs (AdamW, lr\,=\,$10^{-3}$, weight decay\,=\,$10^{-2}$, cosine annealing, Focal Tversky loss, early stopping patience 20). \textbf{FE4} used the self-configuring framework~\cite{isensee2021nnu} for 500 epochs (lr\,=\,$5\times10^{-3}$, weight decay\,=\,$10^{-4}$, 250/50 train/val iterations, 85\% foreground oversampling, deep supervision).

\section{Experiments \& Results}
We evaluated the proposed framework on the SUH cohort ($n=109$) and on ISLES'24 ($n=149$) using BL, GLCM, and CNN embedding features. Both CNN models were used only in inference mode as feature extractors. nnU-Net was used to automatically segment core and penumbra on ISLES'24; on a held-out SUH cohort test set ($n=33$) it achieved Dice scores of $0.71\pm0.24$ (penumbra) and $0.30\pm0.28$ (core), supporting its use for generating the ISLES'24 ROIs. To assess whether admission CTP features at $T_1$ are associated with subsequent tissue fate at $T_2$, we defined four region-pair comparisons between bi-temporal ROI classes. \textbf{Test~1} compares salvaged and infarcted penumbra ($\text{ROI}_{p}^{b}$ vs.\ $\text{ROI}_{p}^{fi}$), evaluating the clinically interesting question of whether admission signatures distinguish penumbral tissue that recovers from tissue that progresses to infarction. \textbf{Test~2} compares salvaged and infarcted core ($\text{ROI}_{c}^{b}$ vs.\ $\text{ROI}_{c}^{fi}$), evaluating whether initially core-labeled tissue retains outcome-related differences. \textbf{Test~3} compares initially non-hypoperfused tissue with subsequent infarction and infarcted penumbra ($\text{ROI}_{NHB}^{fi}$ vs.\ $\text{ROI}_{p}^{fi}$), assessing whether infarcting tissue outside the admission hypoperfused area is separable from penumbra at admission. \textbf{Test~4} compares initially non-hypoperfused tissue with subsequent infarction and healthy contralateral brain ($\text{ROI}_{NHB}^{fi}$ vs.\ $\text{ROI}_{CLB}^{b}$). All comparisons were performed using patient-level feature representations.

\begin{figure}[!htbp]
    \centering
    \includegraphics[width=\linewidth]{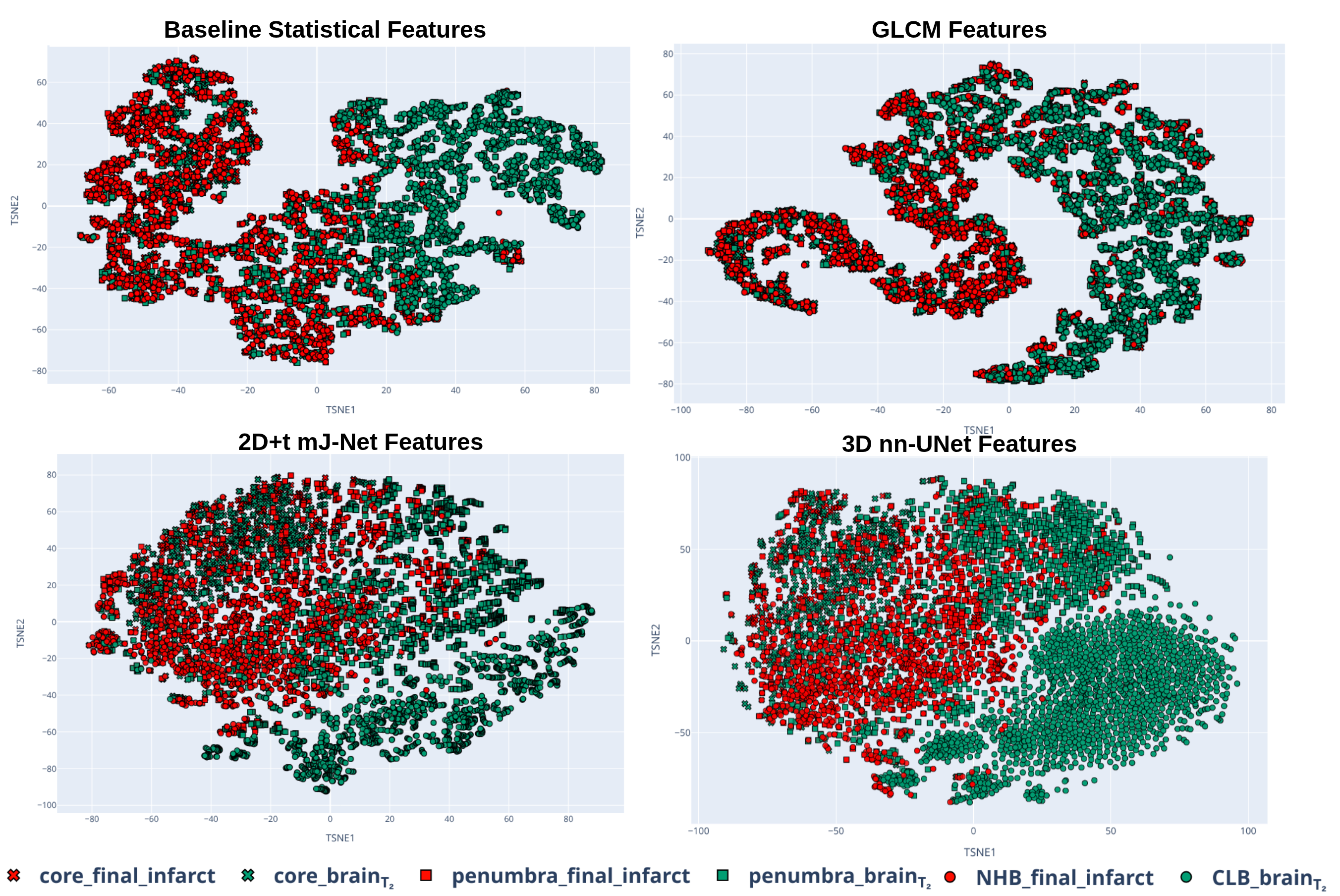}
    \caption{t-SNE projections of FE1 - FE4 across all six bi-temporal ROI classes on the SUH cohort. Colors denote tissue fate and markers denote ROI classes.}
    \label{fig:tsne}
\end{figure}

\begin{figure*}[!htbp]
    \centering
    \includegraphics[width=\linewidth]{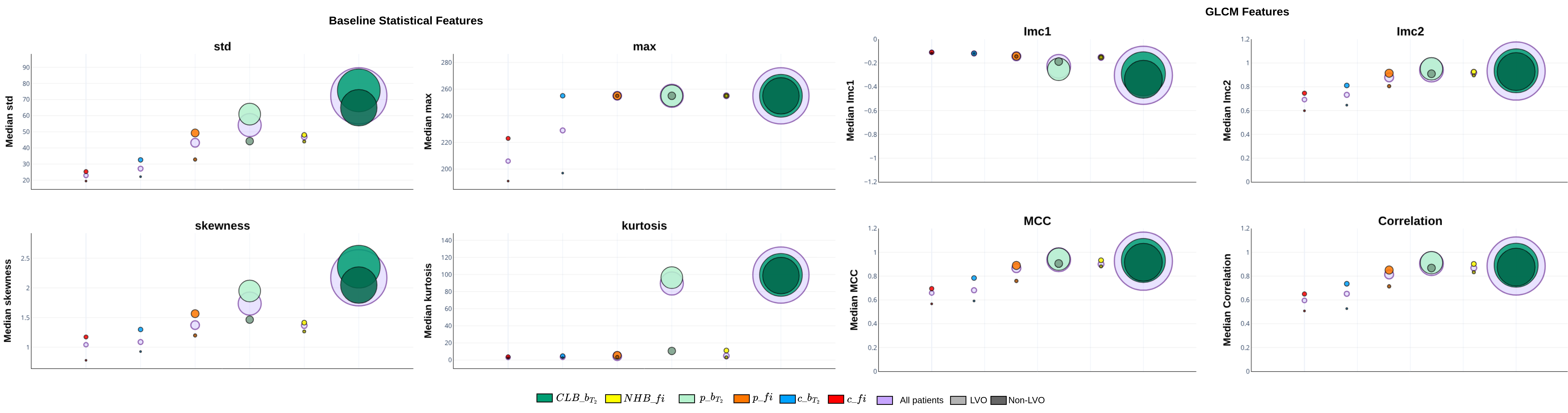}
    \caption{Bubble plots of median BL (Std, Max, Skewness, Kurtosis) and GLCM (Imc1, Imc2, MCC, Correlation) feature values per bi-temporal ROI class on the SUH cohort for LVO (filled circles, larger) and non-LVO (open markers, smaller) patient subgroups. Bubble size encodes voxel count. A consistent phenotypic gradient from healthy ($\text{ROI}_{\textit{CLB}}^{b}$) through salvaged penumbra ($\text{ROI}_{p}^{b}$) to infarcted tissue is visible across both subgroups.}
    \label{fig:bubble}
\end{figure*}

\begin{figure}[htbp]
    \centering
    \includegraphics[width=\linewidth]{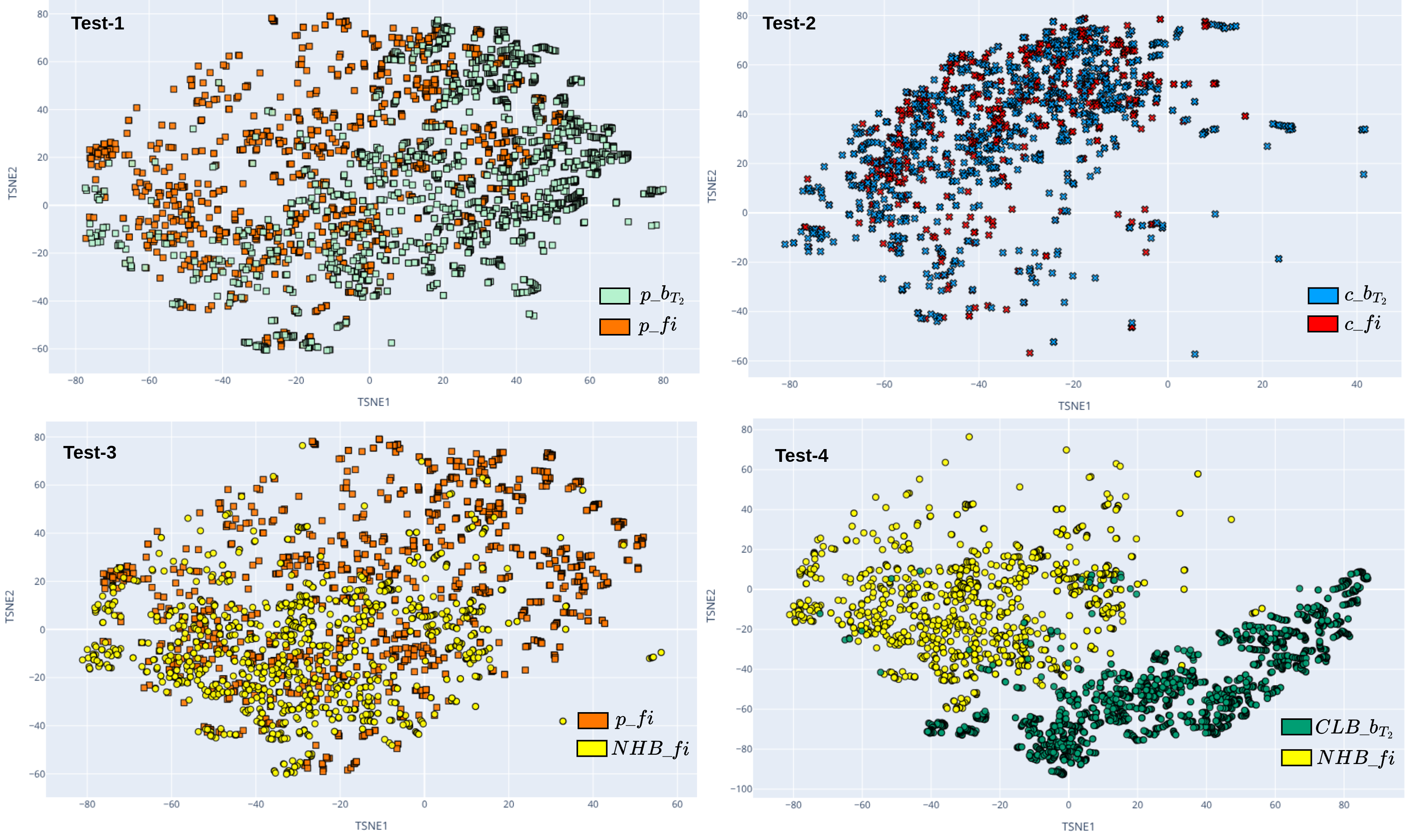}
    \caption{t-SNE projections of FE3 (mJ-Net) embeddings for each region-pair test on the SUH cohort: \textbf{Test~1} ($\text{ROI}_{p}^{b}$ vs $\text{ROI}_{p}^{fi}$), \textbf{Test~2} ($\text{ROI}_{c}^{b}$ vs $\text{ROI}_{c}^{fi}$), \textbf{Test~3} ($\text{ROI}_{\textit{NHB}}^{fi}$ vs $\text{ROI}_{p}^{fi}$), \textbf{Test~4} ($\text{ROI}_{\textit{NHB}}^{fi}$ vs $\text{ROI}_{\textit{CLB}}^{b}$). Colors denote the two ROI classes being compared in each test.}
    \label{fig:statstsne}
\end{figure}

Table~\ref{tab:stats_4tests} summarizes the four region-pair comparisons for BL, GLCM, and CNN embedding features in both cohorts. Fig.~\ref{fig:tsne} presents slice-level t-SNE projections of FE1-FE4 across the six bi-temporal ROI classes. Fig.~\ref{fig:bubble} shows bubble plots of median BL and GLCM feature values per ROI class for the LVO and non-LVO subgroups. Fig.~\ref{fig:statstsne} presents mJ-Net (FE3) t-SNE projections for Tests~1--4 on the SUH cohort. Table~\ref{tab:ablation_norm} reports the ablation study over normalization and pooling aggregation strategies, and Table~\ref{tab:subgroup} presents subgroup analyses across stroke hemisphere and onset-to-CT timing.

\textbf{Ablation study.}
Table~\ref{tab:ablation_norm} evaluates normalization and aggregation choices for CNN embedding separation. Z-score normalization generally improved $\tilde{\Delta}_{\cos}$. Max pooling with Z-score normalization achieved the strongest separation for the primary penumbral fate comparison (Test~1) in both models and was therefore used as a single consistent configuration for the reported CNN results. Although mean aggregation performed better for nnU-Net on Tests~3-4, the max-pooled results provide a conservative estimate for these secondary comparisons.

\begin{table}[!htbp]
\centering
\caption{Ablation: effect of Z-score normalization and pooling aggregation
on CNN embedding separation ($\tilde{\Delta}_{\cos}$, one-sample Wilcoxon, SUH cohort).
All entries significant ($p < 0.05$). \textbf{Bold} = highest per test within model.
}
\label{tab:ablation_norm}
\setlength{\tabcolsep}{4pt}
\renewcommand{\arraystretch}{1.15}
{\fontsize{8}{9}\selectfont
\resizebox{\linewidth}{!}{%
\begin{tabular}{llc |
  >{\centering\arraybackslash}p{1.8cm} |
  >{\centering\arraybackslash}p{1.8cm} |
  >{\centering\arraybackslash}p{1.8cm} |
  >{\centering\arraybackslash}p{1.8cm}}
\toprule
& & & \textbf{Test 1} & \textbf{Test 2} & \textbf{Test 3} & \textbf{Test 4} \\
\textbf{Model} & \textbf{Norm.} & \textbf{Aggr.}
& \makecell{$\text{ROI}_{p}^{b}$\\vs $\text{ROI}_{p}^{fi}$}
& \makecell{$\text{ROI}_{c}^{b}$\\vs $\text{ROI}_{c}^{fi}$}
& \makecell{$\text{ROI}_{NHB}^{fi}$\\vs $\text{ROI}_{p}^{fi}$}
& \makecell{$\text{ROI}_{NHB}^{fi}$\\vs $\text{ROI}_{CLB}^{b}$} \\
\midrule
\multirow{6}{*}{mJ-Net}
& \multirow{3}{*}{None}
  & Mean   & \cellcolor{lightgreen}\checkmark\ 0.068 & \cellcolor{lightgreen}\checkmark\ 0.018 & \cellcolor{medgreen}\checkmark\ 0.209         & \cellcolor{medgreen}\checkmark\ 0.181 \\
& & Median & \cellcolor{lightgreen}\checkmark\ 0.091 & \cellcolor{lightgreen}\checkmark\ 0.023 & \cellcolor{medgreen}\checkmark\ \textbf{0.213} & \cellcolor{medgreen}\checkmark\ 0.215 \\
& & Max    & \cellcolor{lightgreen}\checkmark\ 0.033 & \cellcolor{lightgreen}\checkmark\ 0.015 & \cellcolor{lightgreen}\checkmark\ 0.085        & \cellcolor{medgreen}\checkmark\ 0.119 \\
\cmidrule(lr){2-7}
& \multirow{3}{*}{Z-score}
  & Mean   & \cellcolor{lightgreen}\checkmark\ 0.098        & \cellcolor{lightgreen}\checkmark\ 0.029              & \cellcolor{medgreen}\checkmark\ 0.196         & \cellcolor{medgreen}\checkmark\ 0.264 \\
& & Median & \cellcolor{medgreen}\checkmark\ 0.114           & \cellcolor{lightgreen}\checkmark\ \textbf{0.036}     & \cellcolor{medgreen}\checkmark\ 0.200         & \cellcolor{medgreen}\checkmark\ 0.290 \\
& & Max    & \cellcolor{medgreen}\checkmark\ \textbf{0.146}  & \cellcolor{lightgreen}\checkmark\ 0.031              & \cellcolor{medgreen}\checkmark\ 0.127         & \cellcolor{siggreen}\checkmark\ \textbf{0.460} \\
\midrule
\multirow{6}{*}{nnU-Net}
& \multirow{3}{*}{None}
  & Mean   & \cellcolor{lightgreen}\checkmark\ 0.019 & \cellcolor{lightgreen}\checkmark\ 0.009 & \cellcolor{lightgreen}\checkmark\ 0.061 & \cellcolor{lightgreen}\checkmark\ 0.085 \\
& & Median & \cellcolor{lightgreen}\checkmark\ 0.023 & \cellcolor{lightgreen}\checkmark\ 0.011 & \cellcolor{lightgreen}\checkmark\ 0.074 & \cellcolor{lightgreen}\checkmark\ 0.099 \\
& & Max    & \cellcolor{lightgreen}\checkmark\ 0.011 & \cellcolor{lightgreen}\checkmark\ 0.004 & \cellcolor{lightgreen}\checkmark\ 0.024 & \cellcolor{lightgreen}\checkmark\ 0.051 \\
\cmidrule(lr){2-7}
& \multirow{3}{*}{Z-score}
  & Mean   & \cellcolor{lightgreen}\checkmark\ 0.066         & \cellcolor{lightgreen}\checkmark\ 0.029             & \cellcolor{medgreen}\checkmark\ \textbf{0.140} & \cellcolor{medgreen}\checkmark\ \textbf{0.270} \\
& & Median & \cellcolor{lightgreen}\checkmark\ 0.071         & \cellcolor{lightgreen}\checkmark\ 0.028             & \cellcolor{medgreen}\checkmark\ 0.137          & \cellcolor{medgreen}\checkmark\ 0.259 \\
& & Max    & \cellcolor{lightgreen}\checkmark\ \textbf{0.091} & \cellcolor{lightgreen}\checkmark\ \textbf{0.038}    & \cellcolor{lightgreen}\checkmark\ 0.090        & \cellcolor{medgreen}\checkmark\ 0.193 \\
\bottomrule
\end{tabular}%
}}
\end{table}

\textbf{Subgroup robustness.}
Table~\ref{tab:subgroup} shows that feature separation is consistent across left and right hemisphere strokes, suggesting that the observed separation is not primarily driven by stroke lateralization. Across onset-to-CT timing subgroups (median split: 1.22h), all tests remain significant in both groups. The slightly higher $\tilde{\Delta}_{\cos}$ for early-imaged patients in Test 2 ($0.049$ vs $0.023$) suggests that imaging closer to onset may capture more distinct core tissue phenotypes at admission, though this difference is modest and both groups remain significant.
\begin{table}[t]
\centering
\caption{Subgroup analysis of mJ-Net embedding separation ($\tilde{\Delta}_{\cos}$,
one-sample Wilcoxon, SUH cohort).
\checkmark~= significant; $\times$~= not significant.
}
\label{tab:subgroup}
\setlength{\tabcolsep}{4pt}
\renewcommand{\arraystretch}{1.15}
{\fontsize{8}{9}\selectfont
\resizebox{\linewidth}{!}{%
\begin{tabular}{lc |
  >{\centering\arraybackslash}p{1.8cm} |
  >{\centering\arraybackslash}p{1.8cm} |
  >{\centering\arraybackslash}p{1.8cm} |
  >{\centering\arraybackslash}p{1.8cm}}
\toprule
& & \textbf{Test 1} & \textbf{Test 2} & \textbf{Test 3} & \textbf{Test 4} \\
\textbf{Subgroup} & $n$
& \makecell{$\text{ROI}_{p}^{b}$\\vs $\text{ROI}_{p}^{fi}$}
& \makecell{$\text{ROI}_{c}^{b}$\\vs $\text{ROI}_{c}^{fi}$}
& \makecell{$\text{ROI}_{NHB}^{fi}$\\vs $\text{ROI}_{p}^{fi}$}
& \makecell{$\text{ROI}_{NHB}^{fi}$\\vs $\text{ROI}_{CLB}^{b}$} \\
\midrule
\multicolumn{6}{l}{\textit{Stroke hemisphere}} \\
\midrule
Left  & 53
& \cellcolor{medgreen}\checkmark\ 0.133
& \cellcolor{lightgreen}\checkmark\ 0.028
& \cellcolor{medgreen}\checkmark\ 0.124
& \cellcolor{siggreen}\checkmark\ 0.466 \\
Right & 51
& \cellcolor{medgreen}\checkmark\ 0.155
& \cellcolor{lightgreen}\checkmark\ 0.029
& \cellcolor{medgreen}\checkmark\ 0.143
& \cellcolor{siggreen}\checkmark\ 0.458 \\
\midrule
\multicolumn{6}{l}{\textit{Onset-to-CT time (median split: 1.22h)}} \\
\midrule
$<1.22$h    & 51
& \cellcolor{medgreen}\checkmark\ 0.163
& \cellcolor{lightgreen}\checkmark\ 0.049
& \cellcolor{medgreen}\checkmark\ 0.137
& \cellcolor{siggreen}\checkmark\ 0.453 \\
$\geq1.22$h & 54
& \cellcolor{medgreen}\checkmark\ 0.137
& \cellcolor{lightgreen}\checkmark\ 0.023
& \cellcolor{medgreen}\checkmark\ 0.122
& \cellcolor{siggreen}\checkmark\ 0.448 \\
\bottomrule
\end{tabular}%
}}
{\footnotesize 4 patients excluded due to missing onset-to-CT data.
Bilateral/midline cases excluded from hemisphere analysis.}
\end{table}

\section{Discussion}
Across the four region-pair comparisons (Table~\ref{tab:stats_4tests}), admission CTP features showed a graded organization between tissue-fate classes, suggesting that tissue outcome is reflected in the admission feature space. The most consistent finding was observed in penumbra, where salvaged and infarcted tissue were separated across BL, GLCM, and CNN feature families (Test~1). BL and GLCM features showed mostly medium-to-large effects, while mJ-Net also showed clear embedding separation ($\tilde{\Delta}_{\cos}=0.146$ on the SUH cohort and $0.177$ on ISLES'24). This supports the clinical relevance of penumbral heterogeneity and suggests that admission CTP contains tissue fate associated signatures beyond conventional core-penumbra delineation. In contrast, core tissue showed weak separation by subsequent fate (Test~2), with negligible to small BL/GLCM effects and much smaller mJ-Net separations on SUH and on ISLES'24 ($\tilde{\Delta}_{\cos} = 0.031$ and $0.044$), consistent with advanced ischemic injury at baseline. Tests~3-4 further suggest that tissue outside the admission hypoperfused area but later included in the final infarct may already differ from healthy tissue at admission. This was most evident in Test~4, where both handcrafted features and CNN embeddings showed strong separation, particularly for mJ-Net ($\tilde{\Delta}_{\cos} =0.460$ on SUH and $0.401$ on ISLES'24).

The visual analyses support the same tissue-fate pattern. The t-SNE projections (Figs.~\ref{fig:tsne},~\ref{fig:statstsne}) show clearer class organization in the CNN embedding spaces than in the handcrafted feature spaces, while the bubble plots (Fig.~\ref{fig:bubble}) reveal BL and GLCM phenotypic gradients in which salvaged penumbra lies closer to healthy contralateral brain than to infarcted penumbra, and the two core classes converge. This is clinically plausible: penumbra may either recover or progress to final infarct, whereas core-labeled tissue is expected to be less reversible. Both CNN embeddings showed the clearest separation in the penumbral (Test~1) and Test~4 comparisons, consistent with the handcrafted features.

The same qualitative pattern was observed on ISLES'24 despite differences in centers, acquisition protocols, follow-up interval, and automated rather than manual $T_1$ annotations (Table~\ref{tab:stats_4tests}). Test~1 remained consistently separable and Test~2 remained the weakest comparison across feature families, supporting cross-dataset consistency of the main tissue-fate pattern. Test~3 showed stronger BL/GLCM separation on ISLES'24 than on the SUH cohort, which may partly reflect differences in $T_1$ ROI construction, as automated core-penumbra masks can alter the boundary between infarcted penumbra and initially non-hypoperfused tissue that later infarcts. Overall, the framework is best viewed as a descriptive, hypothesis-generating approach for identifying tissue phenotypes associated with stroke evolution.

\textbf{Limitations.} The CNN extractors were trained on the SUH cohort, so the SUH embedding separations are not fully independent. This is bounded by design: the extractors were trained only for $T_1$ core/penumbra segmentation and never observed the follow-up labels defining tissue fate. Empirically, the training-independent ISLES'24 results and the model-free BL/GLCM features show the same pattern. Bi-temporal ROI construction is also sensitive to residual CTP-DWI misregistration, and the ISLES'24 core/penumbra masks were generated automatically, introducing label uncertainty. Reperfusion status was not modeled: all patients in the SUH analysis cohort received thrombolysis and some thrombectomy, but the reperfusion therapy was not successful in all patients. Reperfusion is one of the most important determinants of penumbral fate, though fast progressors (penumbra infarcting despite rapid reperfusion) and ghost core (presumed core remaining viable) show that additional refinement of the penumbra concept and better understanding of the perfusion parameters are needed. The framework is retrospective and exploratory; larger multi-center studies are needed to confirm the phenotypes and assess clinical utility.

\section{Conclusion and Future Work}
This work presents a bi-temporal tissue phenotyping framework linking admission CTP features to follow-up DWI-defined tissue outcome. Across statistical, radiomic, and CNN representations, penumbra showed consistent retrospective separation by fate while core did not, with CNN embeddings separating most strongly; the same pattern  was observed across SUH clinical subgroups and on the publicly available ISLES'24 cohort. Future work will account for reperfusion status and treatment, construct longitudinal phenotype trajectories, and assess predictive utility in larger multi-center cohorts.

\bibliographystyle{ieeetr}
\bibliography{ieee_bhi_bibliography}
\end{document}